\documentclass[11pt,a4paper]{article}
\usepackage[hyperref]{acl2021}
\usepackage{times}
\usepackage{latexsym}

\usepackage{graphicx}
\usepackage{lipsum}
\usepackage{multicol}
\usepackage{makecell}
\usepackage{array}
\usepackage{amsmath}

\usepackage{subcaption}
\DeclareCaptionSubType*[Alph]{table}
\DeclareCaptionLabelFormat{mystyle}{Table~\bothIfFirst{#1}{ ̃}#2}
\usepackage{microtype}

\usepackage{soul}

\aclfinalcopy 

\title{Counts@IITK at SemEval-2021 Task 8: SciBERT Based Entity And Semantic Relation Extraction For Scientific Data
}

\usepackage{soul}

\author{Akash Gangwar$^{*}$ \qquad    
  Sabhay Jain$^{*}$ \qquad  
  Shubham Sourav\thanks{\quad Authors equally contributed  to this work.} \qquad
  \large{\textbf{Ashutosh Modi}} \\
{Indian Institute of Technology Kanpur (IIT Kanpur)} \\
  {\tt \{akashgnr, sabhayj, ssourav\}@iitk.ac.in}  \\
  {\tt ashutoshm@cse.iitk.ac.in}  \\
}

\date{}

\begin{document}
\maketitle

\begin{abstract}
This paper presents the system for SemEval 2021 Task 8 (MeasEval). MeasEval is a novel span extraction, classification, and relation extraction task focused on finding quantities, attributes of these quantities, and additional information, including the related measured entities, properties, and measurement contexts. Our submitted system, which placed fifth (team rank) on the leaderboard, consisted of SciBERT with [CLS] token embedding and CRF layer on top. We were also placed first in Quantity (tied) and Unit subtasks, second in MeasuredEntity, Modifier and Qualifies subtasks, and third in Qualifier subtask.
\end{abstract}

\section{Introduction}

SemEval 2021 Task 8 (\citealt{MeasEval2021})
is a task for extracting entities and semantic relations between them from a corpus of scientific articles coming from different domains.
Instead of just identifying quantities, the task gives more weightage to parsing and extracting important semantic relations among the extracted entities. This is challenging because texts are ambiguous, and inconsistent, and extraction relies heavily on implicit knowledge. The results of this task can also be used for extractive scientific data summarization. 


Given a scientific text, 
the task is to identify the span of quantities, units, and other attributes of those quantities and related measured entities, properties, and qualifiers, if any. The organizers have divided the task into five subtasks and submissions will be evaluated against all five sub-tasks\footnote{\url{https://competitions.codalab.org/competitions/25770}}.
\begin{enumerate}
    \item \textbf{Quantity Extraction}: For each paragraph of text, identify all Quantity spans.
    \item \textbf{Unit Detection and Modifier Classification}: For each identified Quantity, identify the Unit of measurement and classify additional value Modifiers (count, range, approximate, mean, etc.) that apply to the Quantity.
    \vspace{-3mm}
    \item \textbf{MeasuredEntity and MeasuredProperty Extraction}: For each identified Quantity, identify the MeasuredEntity and MeasuredProperty associated with it.
    \vspace{-3mm}
    \item \textbf{Qualifier Extraction}: Identify and mark the span of any Qualifier that is needed to record additional related context.
    \vspace{-3mm}
    \item \textbf{HasQuantity, HasProperty and Qualifies Extraction}: Identify relationships between Quantity, MeasureEntity, MeasuredProperty, and Qualifier.
\end{enumerate}

\par
\begin{figure}[t]
\centering
\includegraphics[width=1\linewidth]{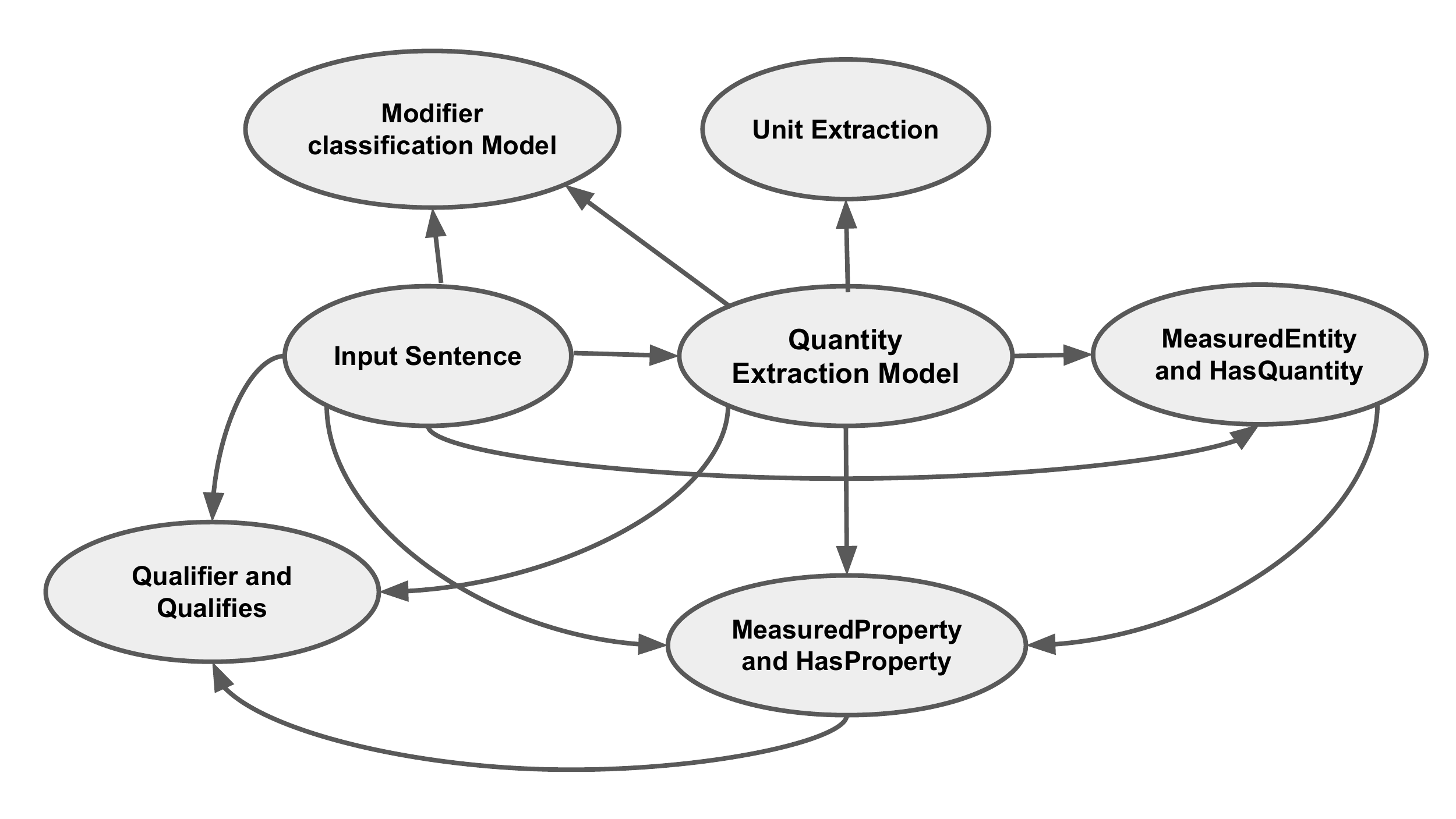}
\vspace{-8mm}
\caption{Overview of our proposed approach} 
\label{pipeline}
\vspace{-4mm}
\end{figure}

 We consider subtask 1 as an entity extraction task, and subtask 3, 4, and 5 
are viewed as relation extraction tasks. 
After extracting the quantities, other attributes (MeasuredEntity, Property, and Qualifier) related to those quantities need to be predicted.
The directed graph in Figure \ref{pipeline} gives an overview of our proposed approach. The set of incoming edges to each node represents the input to the trained model (represented by node), and the label at each node represents the prediction made by the model.
The task data 
is extracted from CC-BY ScienceDirect Articles and made available by the Elsevier Labs 
via the OA-STM-Corpus\footnote{\url{https://github.com/elsevierlabs/OA-STM-Corpus}}.
This motivated the use of SciBERT (\citealt{Beltagy2019SciBERT}) model for various subtasks.  
{``}SciBERT leverages unsupervised pretraining on a large multi-domain corpus of scientific publications to improve performance on downstream scientific NLP tasks{''}. 

\par Our final submitted system consisted of SciBERT with [CLS] token embedding and CRF layer on top, and it achieved an overall F1-overlap score of 0.432. We were ranked fifth on the global leaderboard. The top performance on the leaderboard achieved an overall F1-overlap of 0.519. The implementation of our system is made available via Github\footnote{\url{https://github.com/akashgnr31/Counts-And-Measurement}}.
\par The rest of this paper is arranged as follows. Section 2 introduces the previous work in this field and describes the organizers' dataset. Section 3 explains our overall approach. Section 4 contains the experimental setup for training the model. We conclude with the analysis of our model performance in section 5 and concluding remarks in section 6.

\section{Background}
\subsection{Related work}
\par \citealt{NERLSTM} attempted to recognize the Clinical Entities using  a LSTM CRF based architecture. 
The authors used the word and character level embedding obtained from word2vec  (\citealt{mikolov2013efficient}). For relation extraction between these entities, authors build a binary classifier using random forest classifier. 
This approach has higher time complexity as it checks for all possible relationships that could exist and classifies them.
The more recent work in entity extraction is by \citealt{Lee_2019}, where they fine-tuned the Bert model 
using the Bio-Medical data, and have shown SOTA performance. Some other works in entity extraction includes \citealt{bertPersian}, where they fine-tuned BERT followed by a fully connected layer and a CRF layer.
\par The work by  \citealt{RLBERT2} on Relation Extraction uses BERT to identify the different types of relations between pair of entities in the given text. 
The system does not automatically recognize the entities between which relation exists, rather entities of interest need to be manually specified.  


\subsection{Task setup}
The scientific articles in the training and test corpus are from the following sub-domains: Astronomy, Engineering, Medicine, Materials Science, Biology, Chemistry, Agriculture, Earth Science, and Computer Science. These articles were manually annotated. 
The inter-annotator agreements was calculated using Krippendorff's Alpha IAA score 
(Table \ref{tab:table1}).

\begin{table}[h]
  \begin{center}
    \begin{tabular}{|c|c|}
    \hline
      \textbf{Class} & \textbf{IAA Score}\\
      \hline
      {Quantity} & 0.943\\
      
      {MeasuredEntity} & 0.640\\
     
      {MeasuredProperty} & 0.545 \\
   
      {Qualifier} & 0.333\\
 
      {Units} & 0.866\\
      \hline
    \end{tabular}
    \vspace{-3mm}
  \end{center}
  \caption{IAA scores of various classes}
    \label{tab:table1}
\end{table}
The training dataset comprised of 298 paragraphs containing 1164 quantities, 1148 measured entities, 742 measured properties, and 309 qualifiers. The evaluation set included 135 paragraphs.


\section{System overview}
\subsection{Pre Processing}
Since we are using the SciBERT model, a maximum of 512 tokens can be passed as input to the model. 
Therefore, we used SciSpaCy (\citealt{neumann-etal-2019-scispacy}) to split the paragraph into sentences, and these sentences were passed as input to the SciBERT model.
\subsection{Subtask 1 (Quantity Extraction)}
Input sentences were 
tokenized using a SciBERT tokenizer from HuggingFace (\citealt{wolf-etal-2020-transformers}) implementation. 
The Quantity span 
were transformed into BIO / IOB format
(\citealt{ramshaw-marcus-1995-text}) and used as the true-labels for training the model.


\par The tokenized sentence is passed through SciBERT.
Tanh activation function is applied over the final hidden state of SciBERT i.e. 
\begin{align*}
    H_{i}^{'} = W_{1}[tanh(H_{i})] + b_{1} \;\;\; i = 0, 1, ..., len
\end{align*}
\par 
Here $H_{i}$ is the hidden units corresponding to token $i$ and $len$ is the maximum length of the tokenized sentence.
Similarly, [CLS] token is processed.
\begin{align*}
    H_{cls}^{'} = W_{0}[tanh(H_{0})] + b_{0}
\end{align*}
\par Finally, we get the final representation for the sentence by concatenating $H_{cls}^{'}$ and $H_{i}^{'}$ 
and this is used for prediction via the softmax. 
\begin{gather*}
    H_{i}^{''} = W_{2}[concat(H_{i}^{'}, H_{cls}^{'})] + b_{2} \;\;\; i = 0, 1, ..., len \\
    H^{''} = \left[H_{0}^{''}, H_{1}^{''}, ......, H_{len}^{''} \right]^{T}\\
    p = softmax(H^{''}, \; dim = -1)
\end{gather*}
\par Matrices $W_{0}$ and $W_{1}$ have same dimension, i.e., $W_{0} \in R^{d \times d}, W_{1} \in R^{d \times d}, W_{2} \in R^{t \times 2d}$, where $d$ is the hidden state size from BERT and $t$ represent the number of tags, i.e., $t$ = 3 in our case as we are using BIO encoding.. 

\par CRF (Conditional Random Field)
(\citealt{laffertyCrf})  is a probabilistic model that makes it possible to extract structural dependencies among the BIO tags. The tag probability vector for all the tokens, i.e., $p$, is passed through the CRF layer to generate the most probable output sequence. 
\par We trained the model using CRF loss and Adam optimizer. The overall architecture of the model is shown in Figure \ref{SciBERT_CRF}. The tuned hyper-parameters are reported in appendix \ref{appendix:a}. 
\begin{figure*}[h]
  \includegraphics[width=\textwidth]{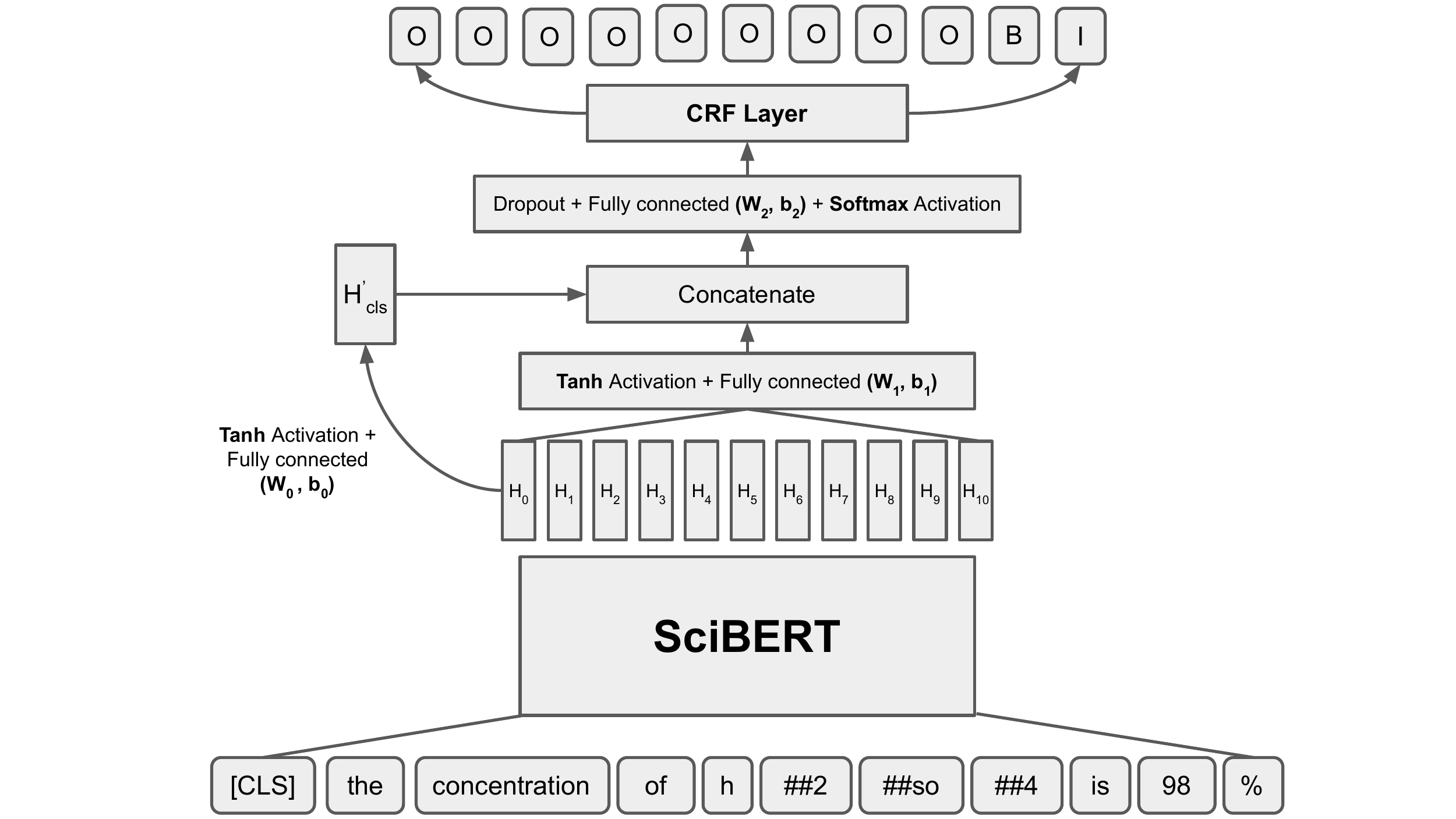}
  \caption{SciBERT with [CLS] token embedding and CRF layer on top (SciBERT + CRF Model)}
  \label{SciBERT_CRF}
\end{figure*}

\subsection{Subtask 2 (Unit Detection)} 
The Quantity phrases are tokenized using Spacy 
(\citealt{spacy}) character-based tokenizer. The true-label for training is formatted as a binary vector marking one at the 
indices for characters in the unit's span in the Quantity phrases. 
\par We trained a Character-based Bi-LSTM (\citealt{10.1162/neco.1997.9.8.1735}) model with trainable word embeddings using BCE (Binary Cross Entropy) loss and Adam optimizer. The model architecture and tuned hyper-parameters are reported in appendix \ref{appendix:b}

\subsection{Subtask 2 (Modifier Classification)}
We formulated this subtask as a multi-label classification problem with 12 labels (HasTolerance, IsApproximate,  IsCount, IsList, IsMean, IsMeanHasSD, IsMeanHasTolerance, IsMeanIsRange, IsMedian, IsRange, IsRangeHasTolerance, None). To 
enable the BERT module to 
capture the location 
of a quantity, we insert the special symbol ``\$" 
at the beginning and end of the Quantity span. If there are multiple Quantities in a sentence, multiple copies of the same sentence are generated with ``\$" at different positions. Suppose $H_{i}$ to $H_{j}$ are the final hidden state vector for the Quantity span. Then, the average operation is applied to get the vector representation of the Quantity. The averaged output is passed through a fully connected layer followed by softmax activation.
\begin{gather*}
    H_{q}^{'} = W\left[tanh\left(\frac{1}{j-i+1}\sum_{k = i}^{j}H_{k}\right)\right] + b\\
    p_{q} = sigmoid(H_{q}^{'})
\end{gather*}

\par Matrix W has dimension $R^{l \times d}$, where $l$ represnts the number of classification label, i.e., $l$ = 12 in our case and $d$ is the hidden state size from BERT.
\par The above model was trained using BCE (Binary Cross Entropy) and Adam optimizer. The threshold value for prediction was determined using cross-validation. The model architecture and tuned hyper-parameters are reported in appendix \ref{appendix:c}.

\subsection{Subtask 3 and 5 (MeasuredEntity and HasQuantity Extraction)}

As done in the previous subtask to capture the location, we insert the special symbol ``\$" at the beginning and end of the quantity span. 
The modified sentences are tokenized using a SciBERT tokenizer. The span of the MeasuredEntity related to Quantity enclosed in the ``\$" symbol is transformed into BIO / IOB format and used as the true-label for training the model.
\par The formatted data is used to train a model similar to the Quantity Extraction (SciBERT + CRF Model). The above model extracts the MeasuredEntity associated with the Quantity enclosed in ``\$". Thus, it predicts the MeasuredEntity as well as the HasQuantity relationship of the predicted MeasuredEntity.

\subsection{Subtask 3 and 5 (MeasuredProperty and HasProperty Extraction)}
To extract MeasuredProperty and HasProperty relationship, we used a similar approach as used for MeasuredEntity and HasQuantity. We enclosed the Quantity span in ``\$" symbol and the MeasuredEntity span in ``\#" symbol.
The modified sentences are passed through the SciBERT tokenizer. The span of MeasuredProperty related to MeasuredEntity, Quantity pair is transformed into BIO / IOB format and used as the true-label for training the model.
\par The formatted data is used to train a model similar to the Quantity Extraction (SciBERT + CRF Model). The model trained is used to extract MeasuredProperty linked with the MeasuredEntity, Quantity pair. If the above model predicts Measured-Property's span, then the HasQuantity relation is updated to MeasuredProperty, and the HasProperty relation is added to MeasuredEntity.

\subsection{Subtask 4 and 5 (Qualifier and Qualifies Extraction)}
To extract Qualifier and Qualifies's span, two separate models similar to Quantity Extraction (SciBERT + CRF Model) were trained. While training the first model, we insert ``\$" at the beginning and end of the Quantity span because we assumed that Qualifier Qualifies Quantity. During the second model training, we enclosed the MeasuredProperty span in ``\$" because of the assumption that Qualifier Qualifies MeasuredProperty. 


\subsection{Post Processing}
Once the predictions from all the models are available, we need to transform the predicted BIO/ IOB format into entity span format. We initially map each token's span in the tokenized sentence and use it to determine the predicted entity's span. While finding the span of the MeasuredEntity, MeasuredProperty, or Qualifier, if our model predicts multiple entities, then we predict the one which is closest to the Quantity span. After that, we convert the sentence span of each entity extracted to the paragraph span.

\section{Experimental Setup}
\begin{table*}
\begin{subtable}{\textwidth}
  \begin{center}
    \begin{tabular}{|c|c|c|c|c|c|c|}
    \hline
      \textbf{Model}  & \textbf{Data Set} & \textbf{Quantity} &
      \textbf{Unit} & \textbf{Modifier} &
      \textbf{MeasuredEntity}& \textbf{MeasuredProperty} \\
      
      \hline
      SciBERT + CRF & eval & 0.861 & 0.804 & 0.614 & 0.406 & 0.245 \\
      \hline
      SciBERT + CRF & dev & 0.887 & 0.744 & 0.696 & 0.322 & 0.216 \\
      \hline
      BERT-Med. + CRF & eval & 0.791 & 0.675 & 0.379 & 0.302 & 0.163 \\
      \hline
    \end{tabular}
    
  \subcaption{}
  
    \label{tab:table2A}
    \vspace*{0.5mm}
    \end{center}
    \end{subtable}
    \begin{subtable}{\textwidth}
    \begin{center}

    \begin{tabular}{|c|c|c|c|c|m{1.8cm}|m{2cm}|}
    \hline
      \textbf{Model}  & \textbf{Data Set} & \textbf{Qualifier} & \textbf{HasQuantity}& \textbf{HasProperty} & \makecell{\centering \textbf{Qualifies}} & \makecell{\textbf{Overall F1} \\ \textbf{overlap}}\\
      \hline
      SciBERT + CRF & eval & 0.077 & 0.311 & 0.183 & \makecell{\centering 0.064} & \makecell{\centering \textbf{0.432}} \\
      \hline
      SciBERT + CRF & dev & 0.083 & 0.270 & 0.137 & \makecell{\centering 0.083} & \makecell{\centering 0.410} \\
      \hline
      BERT-Med. + CRF & eval & 0.0 & 0.193 & 0.114 & \makecell{\centering 0.0} & \makecell{\centering 0.330} \\
      \hline
    \end{tabular}
    \subcaption{}
    \label{tab:table2B}
  \end{center}
  
  \end{subtable}
  \begin{center}
      
  \end{center}
  \vspace*{-0.5cm}
  \caption{Table \ref{tab:table2A} represents the F1-overlap score for subtask 1, 2, 3, and Table \ref{tab:table2B} represents the F1-overlap score for subtask 4, 5 and overall F1-overlap 
  }
    \label{tab:table2}
\end{table*}

The dataset is split into two parts - train set and dev set in a ratio of 90:10. The models were trained on the train set and were validated on the dev set. The environment and packages used for training and pre-processing are listed in appendix \ref{appendix:d}.

\subsection{Evaluation Metrics}
The official metrics used by the SemEval organizer are F1-measure, F1-overlap, and Exact Match. Exact Match is a binary value of 0 or 1, while F1-measure is a token level overlap ratio of submission to true spans, where tokenization is done using simple white space delimiters. F1-overlap is a SQuAD \cite{rajpurkar2016squad} style Overlap score based on F1-measure, which penalizes the negative submissions more strictly. The final evaluation is based on a global F1-overlap score averaged across all subtasks.

\section{Results}

\subsection{Model Variants Used}
We tried various models like BERT-Base, BERT-Medium \cite{devlin2018bert}, SciBERT, and BioBERT (\citealt{Lee_2019}). We could not try BERT-Large due to computational limitations. The results for the top two models are shown in Table \ref{tab:table2}.

We also experimented with Bi-LSTM layers on top of BERT, but the model was overfitting due to its high complexity. Consequently, it was not included in the final model. 

\subsection{Results on evaluation set}
\begin{figure}[!ht]
\centering
\includegraphics[width=1\linewidth]{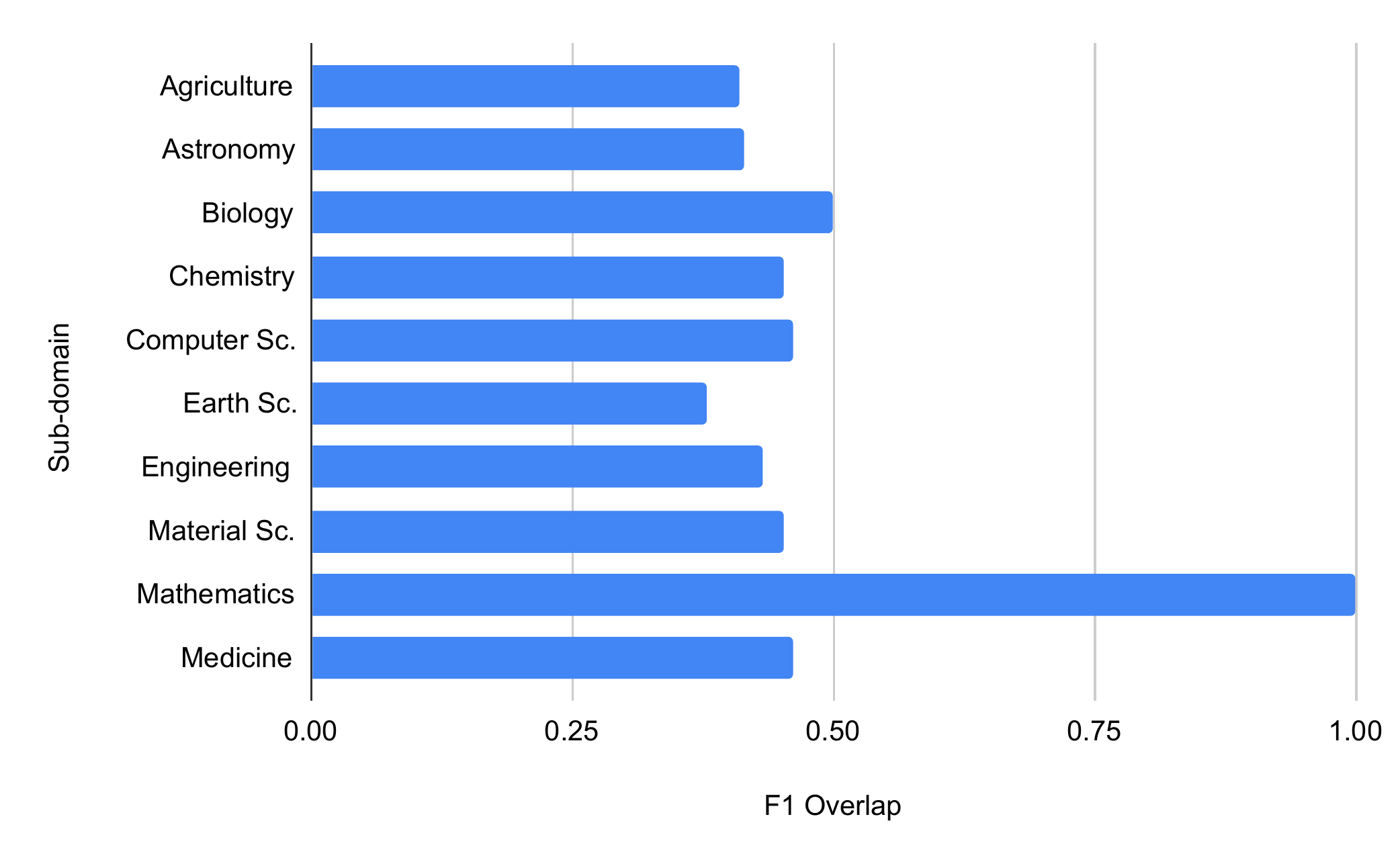}
\vspace{-8mm}
\caption{F1-Overlap scores of various sub-domains} 
\label{fig:1iteration}
\end{figure}

The results achieved on the evaluation set for each subtask are shown in Table \ref{tab:table2}, and the overall results are shown in Table \ref{tab:table3}. 
Figure \ref{fig:1iteration} represents the results achieved in the various subdomains. 

\begin{table}
  \begin{center}
    \begin{tabular}{|c|c|c|}
    \hline
      \textbf{Metric}  & \textbf{SciBERT + CRF} & \textbf{Base line}\\
      \hline
      {Precision} & 0.703 & -\\
      
      {Recall} & 0.560 & -\\
     
      {F-Measure} & 0.623 & -\\
      
      {F1-overlap} & 0.432 & 0.239\\
      
      {Exact Match} & 0.371 & 0.211\\
      \hline
    \end{tabular}
    \vspace{-3mm}
  \end{center}
  \caption{Overall Results on evaluation set}
    \label{tab:table3}
\end{table}

The difference between the Exact Match score and F1-overlap score  shows that the spans predicted by our model were precisely the same as gold data whenever they matched.  

We achieved an overall fifth rank (among 19 participating teams) in the competition. We were also placed first in Quantity (tied) and Unit subtasks, second in MeasuredEntity, Modifier and Qualifies subtasks, and third in Qualifier subtask.

\subsection{Error Analysis}
The relation extraction subtask was challenging because associating entities with the quantities they are related to is context-dependent and based on one's understanding. This is also evident from the IAA scores reported for the train data that even humans can achieve deficient performance.
\par Some of the aspects where our model did not work well are: 
\begin{enumerate}
\item Our model looks for relations only within a sentence, which may cause problems when a relation exists outside the same sentence.
\vspace{-0.75em}
\item There is loss in reconstructing the TSV files from entities because the neighboring data may/maynot be part of the same entity group
\vspace{-0.75em}
\item Our model didn't work well on MeasuredProperty and Qualifiers as it did on other subtasks, which is evident as we achieved only 0.53 and 0.35 F1-overlap on training data for these two subtasks.

\end{enumerate}

\section{Conclusion}
This paper proposed SciBERT + CRF Model (SciBERT with [CLS] token embedding and CRF layer on top) for span extraction, classification, and semantic relation extraction. Our model shows significant improvement in performance over the baseline model and works equally well across all the scientific sub-domains. In the future, we plan to explore various other pre-trained contextual models for our approach.



\bibliographystyle{acl_natbib}
\bibliography{anthology,acl2021}
\newpage
\section*{Appendix}
\appendix
\section{Model Training}
\subsection{SciBERT + CRF}
\label{appendix:a}
In this section we provide hyper-parameter values (Table \ref{tab:table4}) we used for training our final model to facilitate reproduciblity of our results.
\begin{table}[h]
  \begin{center}
    \begin{tabular}{|c|c|}
    \hline
      \textbf{Hyper-parameters} & \textbf{Value}\\
      \hline
      {Hidden State Dimension ($d$)} & 768\\
      
      {Number of tags ($t$)} & 3\\
      
      {Dropout} & 0.1 \\
      
      {Batch Size} & 24\\
     
      {Max Length ($len$)} & 255\\
      
      {Learning Rate} & $10^{-5}$\\
      \hline
    \end{tabular}
    \vspace{-3mm}
  \end{center}
  \caption{Hyper-parameters}
  \label{tab:table4}
\end{table}
\subsection{Unit Detection (Character-based Bi-LSTM)}
\label{appendix:b}
\begin{figure}[h]
\centering
\includegraphics[trim={6cm 2.5cm 5cm 3cm},clip, width=1\linewidth]{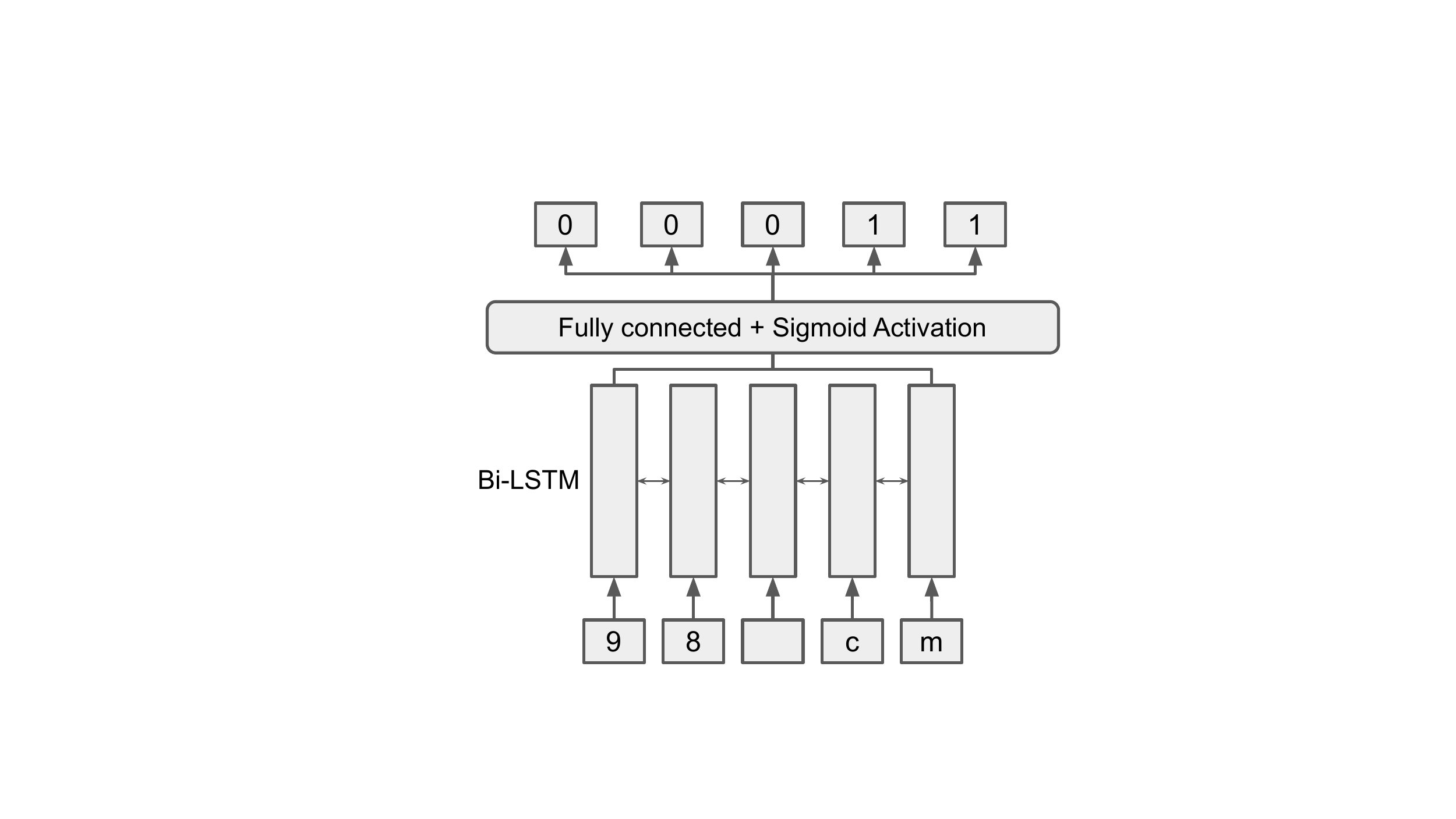}
\vspace{-8mm}
\caption{Character-based Bi-LSTM} 
\label{Bi_LSTM}
\vspace{-4mm}
\end{figure}

In this section we provide model architecture (Figure \ref{Bi_LSTM}) and  hyper-parameter values (Table \ref{tab:table5}) we used for training our final unit extraction model to facilitate reproduciblity of our results.
\begin{table}[h]
  \begin{center}
    \begin{tabular}{|c|c|}
    \hline
      \textbf{Hyper-parameters} & \textbf{Value}\\
      \hline
      {Hidden State Dimension of Bi-LSTM} & 32\\
      
      {Number of Bi-LSTM layers} & 1\\
      
      {Batch Size} & 38\\
     
      {Max Length ($len$)} & 64\\
      
      {Learning Rate} & $10^{-4}$\\
      \hline
    \end{tabular}
    \vspace{-3mm}
  \end{center}
  \caption{Hyper-parameters}
  \label{tab:table5}
\end{table}

\subsection{Modifier Classification (SciBERT with embedding averaging)}
\label{appendix:c}
\begin{figure}[h]
\centering
\includegraphics[trim={4cm 0cm 4cm 0cm}, clip, width=1\linewidth]{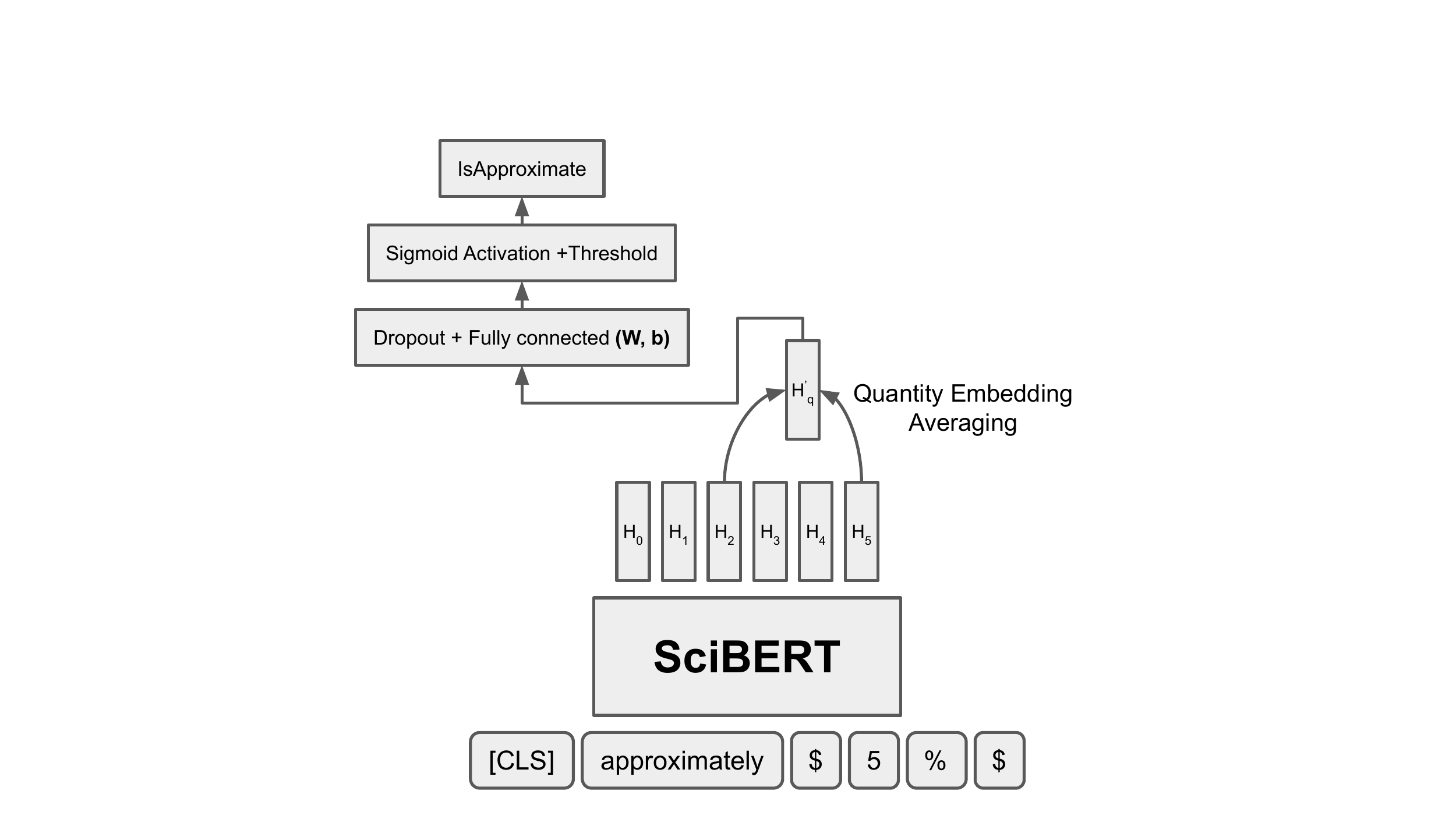}
\vspace{-8mm}
\caption{SciBERT with embedding averaging} 
\label{SciBERT_em_avg}
\end{figure}

In this section we provide model architecture (Figure \ref{SciBERT_em_avg}) and hyper-parameter values (Table \ref{tab:table6}) we used for training our modifier classification final model to facilitate reproduciblity of our results.
\begin{table}[h]
  \begin{center}
    \begin{tabular}{|c|c|}
    \hline
      \textbf{Hyper-parameters} & \textbf{Value}\\
      \hline
      {Hidden State Dimension ($d$)} & 768\\
      
      {Number of labels ($l$)} & 12 \\
      
      {Dropout} & 0.1 \\
      
      {Batch Size} & 24\\
     
      {Max Length ($len$)} & 255\\
      
      {Learning Rate} & $10^{-5}$\\
      
      {Threshold} & 0.5\\
      \hline
    \end{tabular}
    \vspace{-3mm}
  \end{center}
  \caption{Hyper-parameters}
  \label{tab:table6}
\end{table}

\section{Tools/Libraries used}
\label{appendix:d}
We used Google Colab Nvidia T4 GPU (16GB) for training purpose. Python packages (alongwith version) used for pre-processing and training are tabulated below:
\begin{table}[h]
  \begin{center}
    \begin{tabular}{|c|c|}
    \hline
      \textbf{Package} & \textbf{Version}\\
      \hline
      {transformers} & 4.3.2\\
      
      {torchcrf} & 0.7.2\\
      
      {torch} & 1.7.0 \\
      
      {scikit-learn} & 0.22.2\\
      
      {en\_core\_sci\_sm} & 0.3.0\\
      
      {Stanza} & 1.2\\
      
      {spaCy} & 2.3.5\\
      
      {NLTK} & 3.2.5\\
    
      {pandas} & 1.1.5\\
    
      {NumPy} & 1.19.5\\
      \hline
    \end{tabular}
    \vspace{-3mm}
  \end{center}
  \caption{Python Packages}
\end{table}

\end{document}